# When BERT Fails – The Limits of EHR Classification

Augusto Garcia-Agundez, PhD[1], Carsten Eickhoff, PhD[1]
[1]Brown University, Providence, RI, USA

**Abstract**
*Transformers are powerful text representation learners, useful for all kinds of clinical decision support tasks. Although they outperform baselines on readmission prediction, they are not infallible. Here, we look into one such failure case, and report patterns that lead to inferior predictive performance.*

**Introduction**
Transformers such as BERT have shown great potential for clinical decision support (e.g. readmission prediction), often outperforming baselines[1]. However, the improvement they bring is inferior to the one reported in general domain Natural Language Processing (NLP). In this work, we explore the predictive errors of one such task: predicting death of subendocardial infarction patients using early notes to construct a time series of contextual embeddings.

**Materials and Methods**
Our cohort (n=1362), obtained from MIMIC-III[2], consists of patients presenting with subendocardial infarction (ICD9 410.71) with a length of stay (LOS) above the cohort median of 8 days. We select this cohort for its size and mortality, and the LOS cutoff to balance available data and number of excluded patients. The number of included notes per category is: 1649 echocardiography, 6803 ECG, 6353 nursing, 9691 radiology and 12683 nursing/other. We perform a binary classification of outcomes as survival (n=1184) or death (n=178) by constructing a day-resolution time series: We create a text bucket per patient, note category, and day and extract the pretrained ClinicalBERT[1] CLS token contextual embeddings. If a given bucket is empty (e.g. no ECG content on a given day) the bucket of the previous day is copied. If the first day is empty, it is left empty. We then train *n* single-layer classification heads concatenating CLS contextual embeddings, using data for *n* days, with *n* from 1 (only notes from the day of admission) to 8 (all notes up to 8 days). Each task is repeated 100 times, with a 25% train-test split, 100 epochs and a learning rate of 1e-5. The train set is balanced to 1:1 to minimize the impact of class imbalance. We report the aggregated results of all test sets, resulting in approximately 34,000 predictions for each set of included days.

**Results**
Table 1 presents the confusion matrix at days 1, 2, 4 and 8. While the classifier becomes more pessimistic as days pass, accuracy barely changes. An analysis of error distribution shows more insight into how this happens: Figure 1 depicts a histogram of patients, divided by their true class and predictive accuracy. The y axis represents the percentage for their class (% of survivors and mortalities), while the x axis depicts their percentage of correct predictions (e.g. the leftmost bars depict patients for which predictions were correct less than 10% of the time). This reveals a group of approximately 25% of survivors (318) in which classification is nearly always incorrect, as the algorithm predicts death with very high confidence. Compared to the rest of the cohort, this group is 10 times more often a readmission, shows 50% more readmissions in the future, 8 times as many readmissions within the next 30 days, more numerous and semantically distinct nursing notes, especially later in the stay, and a 35% longer LOS.

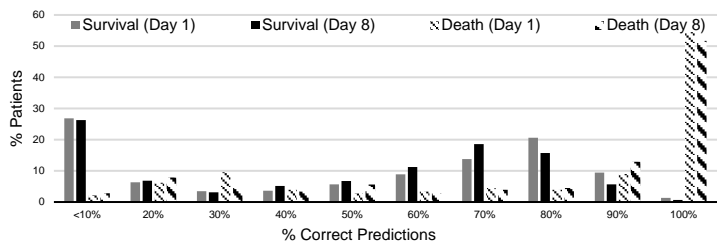

**Figure 1.** Histogram of correct predictions at admission and median LOS for survivals and deaths

|  |  | True Class | |
|---|---|---|---|
|  |  | Survival | Death |
| Predicted Class Day 1 | Survival | 13394 | 1032 |
|  | Death | 16206 | 3468 |
| Predicted Class Day 2 | Survival | 13674 | 1067 |
|  | Death | 15926 | 3433 |
| Predicted Class Day 4 | Survival | 13301 | 981 |
|  | Death | 16299 | 3519 |
| Predicted Class Day 8 | Survival | 12692 | 1009 |
|  | Death | 16908 | 3491 |

**Table 1.** Confusion matrix

**Conclusion**
The overlap between incorrect, high confidence death predictions and readmissions suggests the presence of confounders in the notes for this classification task. In the future, we will corroborate our findings with a larger dataset, identify these potential confounders, and study the impact of each individual note within a time series.

**Acknowledgements**
This study was supported by the Marie Skłodowska-Curie Action MAESTRO, Grant Number 101027770